%% file: main.tex
\documentclass[conference]{IEEEtran}
\usepackage{blindtext, graphicx}
\usepackage{subcaption}
\usepackage{amsmath}
\usepackage{amssymb}
\usepackage{color}
\usepackage{colortbl}
\usepackage{graphicx}
\usepackage{acronym}
\ifCLASSINFOpdf
\else
\fi

\acrodef{lstm}		[\textsc{LSTM\xspace}]				{Long Short-Term Memory Neural Networks}
\acrodef{rnn}		[\textsc{RNN\xspace}]				{Recurrent Neural Networks}
\acrodef{dip}		[\textsc{DIP\xspace}]				{Document Image Processing Pipeline}
\acrodef{dipp}		[\textsc{DIPP\xspace}]				{Document Image Processing Pipeline}

\acrodef{cnn}		[\textsc{CNN\xspace}]				{Convolutional Neural Networks}

\acrodef{relu}		[\textsc{ReLUs\xspace}]				{Rectified Linear Units}

\usepackage[noadjust]{cite}

\acrodef{das}[\textsc{DAS}\xspace]{Document Analysis System}
\acrodef{ann}[\textsc{ANN}\xspace]{Artificial Neural Network}
\acrodef{cnn}[\textsc{CNN}\xspace]{Convolutional Neural Network}
\acrodef{fcn}[\textsc{FCN}\xspace]{Fully Connected Network}
\acrodef{gpu}[\textsc{GPU}\xspace]{Graphics Processing Unit}
\acrodef{svm}[\textsc{SVM\xspace}]{Support Vector Machine}
\acrodef{pca}[\textsc{PCA\xspace}]{Principal Component Analysis}
\acrodef{cc}[\textsc{CC\xspace}]{Connected Component}
\acrodef{bvlc}[\textsc{BVLC\xspace}]{Berkeley Vision and Learning Center}
\acrodef{sf}[\textit{smart FIX}]{\textit{smart \textbf{F}or \textbf{I}nformation e\textbf{X}traction}}
\acrodef{dfki}[\textsc{DFKI\xspace}]{German Research Center for Artificial Intelligence}
\acrodef{gt}[\textsc{GT}\xspace]{Ground Truth}
\acrodef{fp}[\textsc{$FP$}\xspace]{False Positives}
\acrodef{fn}[\textsc{$FN$\xspace}]{False Negatives}	
\acrodef{tp}[\textsc{$TP$\xspace}]{True Positives}
\acrodef{tn}[\textsc{$TN$\xspace}]{True Negatives}
\acrodef{ocr}[\textsc{OCR\xspace}]{Optical Character Recognition}
\acrodef{pkv}[\textsc{PKV\xspace}]{Private Health Insurance}

\acrodef{slfn}[\textsc{SLFN\xspace}]{Single Layer Feedforward Network}
\acrodef{elm}[\textsc{ELM\xspace}]{Extreme Learning Machines}

\newcommand*{\eg}		{e.g.}
\newcommand*{\ie}		{i.e.}

\newcommand*{\cf}		{cf.}
\newcommand*{\sota}		{state-of-the-art}

\begin{document}

\title{Real-Time Document Image Classification using Deep CNN and Extreme Learning Machines}


\input{authors.tex}

\maketitle

\IEEEpeerreviewmaketitle
\input{abstract.tex}

\input{introduction.tex}

\input{relatedwork.tex}

\input{elm.tex}

\input{deepcnn.tex}

\input{experiments.tex}

\input{conclusion.tex}

\ifCLASSOPTIONcaptionsoff
  \newpage
\fi

\bibliographystyle{IEEEtran}
\bibliography{sample.bib}

\end{document}

%% file: authors.tex

\author{
    \IEEEauthorblockN{ Andreas K\"olsch\IEEEauthorrefmark{1}\IEEEauthorrefmark{2},
    Muhammad Zeshan Afzal\IEEEauthorrefmark{1}\IEEEauthorrefmark{2},
    Markus Ebbecke\IEEEauthorrefmark{2},
    Marcus Liwicki\IEEEauthorrefmark{1}\IEEEauthorrefmark{2}\IEEEauthorrefmark{3}}
    
    \IEEEauthorblockA{
            a\_koelsch12@cs.uni-kl.de,
            afzal@iupr.com,
            m.ebbecke@insiders-technologies.de,
            marcus.liwicki@unifr.ch
    }
    
    \vspace{3pt}
    
    \IEEEauthorblockA{\IEEEauthorrefmark{1}MindGarage, University of Kaiserslautern, Germany}

    \IEEEauthorblockA{\IEEEauthorrefmark{2}Insiders Technologies GmbH, Kaiserslautern, Germany}
    
    \IEEEauthorblockA{\IEEEauthorrefmark{3}University of Fribourg, Switzerland}   
    
}

%% file: abstract.tex
\begin{abstract}
This paper presents an approach for real-time training and testing for document image classification. In production environments, it is crucial to perform accurate and (time-)efficient training. 
Existing deep learning approaches for classifying documents do not meet these requirements, as they require much time for training and fine-tuning the deep architectures.
Motivated from Computer Vision, we propose a two-stage approach. The first stage trains a deep network that works as feature extractor and in the second stage, Extreme Learning Machines (ELMs) are used for classification.
The proposed approach outperforms all previously reported structural and deep learning based methods with a final accuracy of $83.24\,\%$ on Tobacco-3482 dataset, leading to a relative error reduction of $25\,\%$ when compared to a previous \ac{cnn} based approach (DeepDocClassifier).
More importantly, the training time of the ELM is only $1.176$ seconds and the overall prediction time for $2,482$ images is $3.066$ seconds.
As such, this novel approach makes deep learning-based document classification suitable for large-scale real-time applications.


\end{abstract}



%% file: introduction.tex
\section{Introduction}

\begin{figure}
    \centering
        \fbox{\includegraphics[width=0.24\linewidth,height=0.12\textheight]{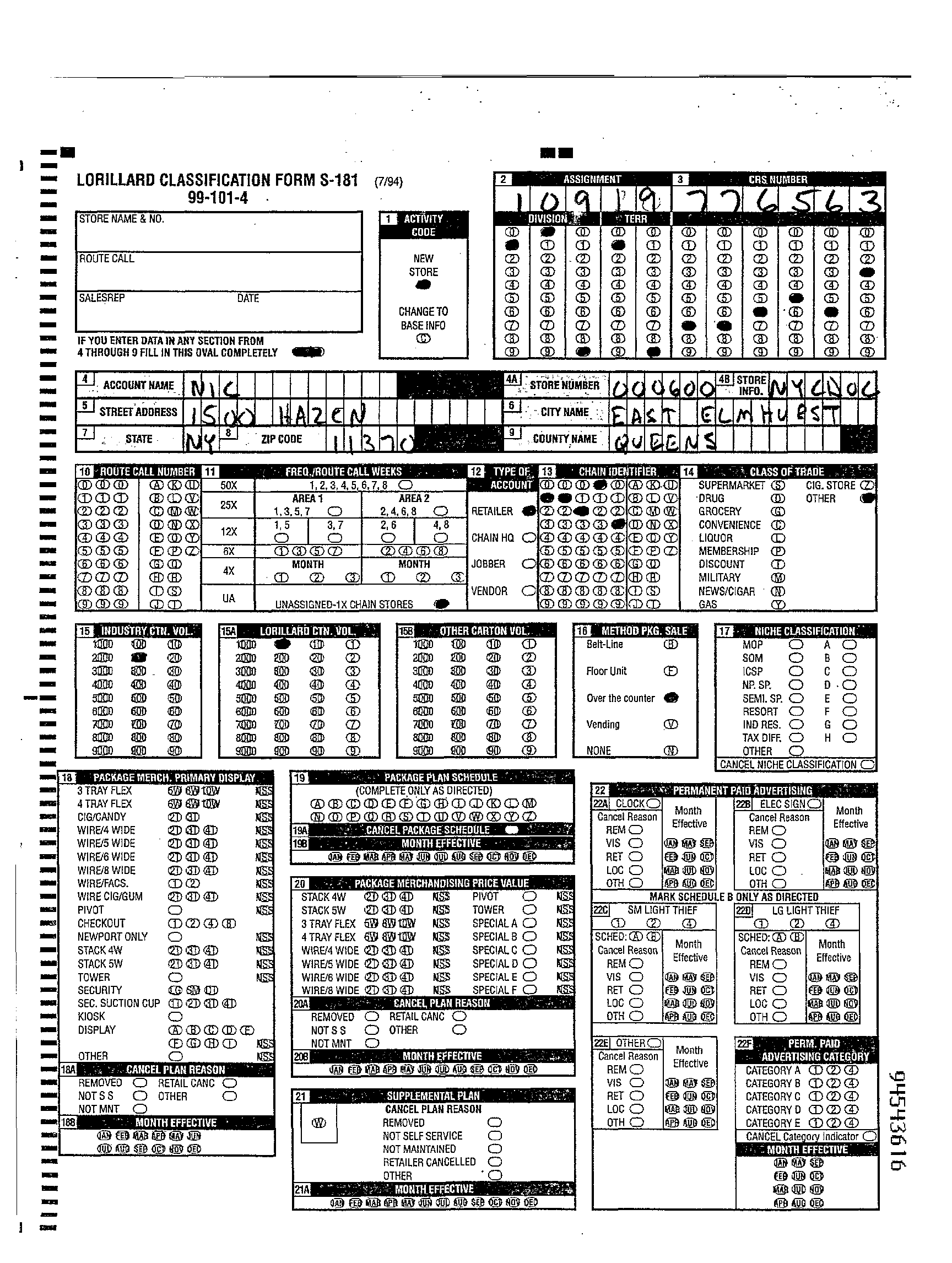}}
        \fbox{\includegraphics[width=0.24\linewidth,height=0.12\textheight]{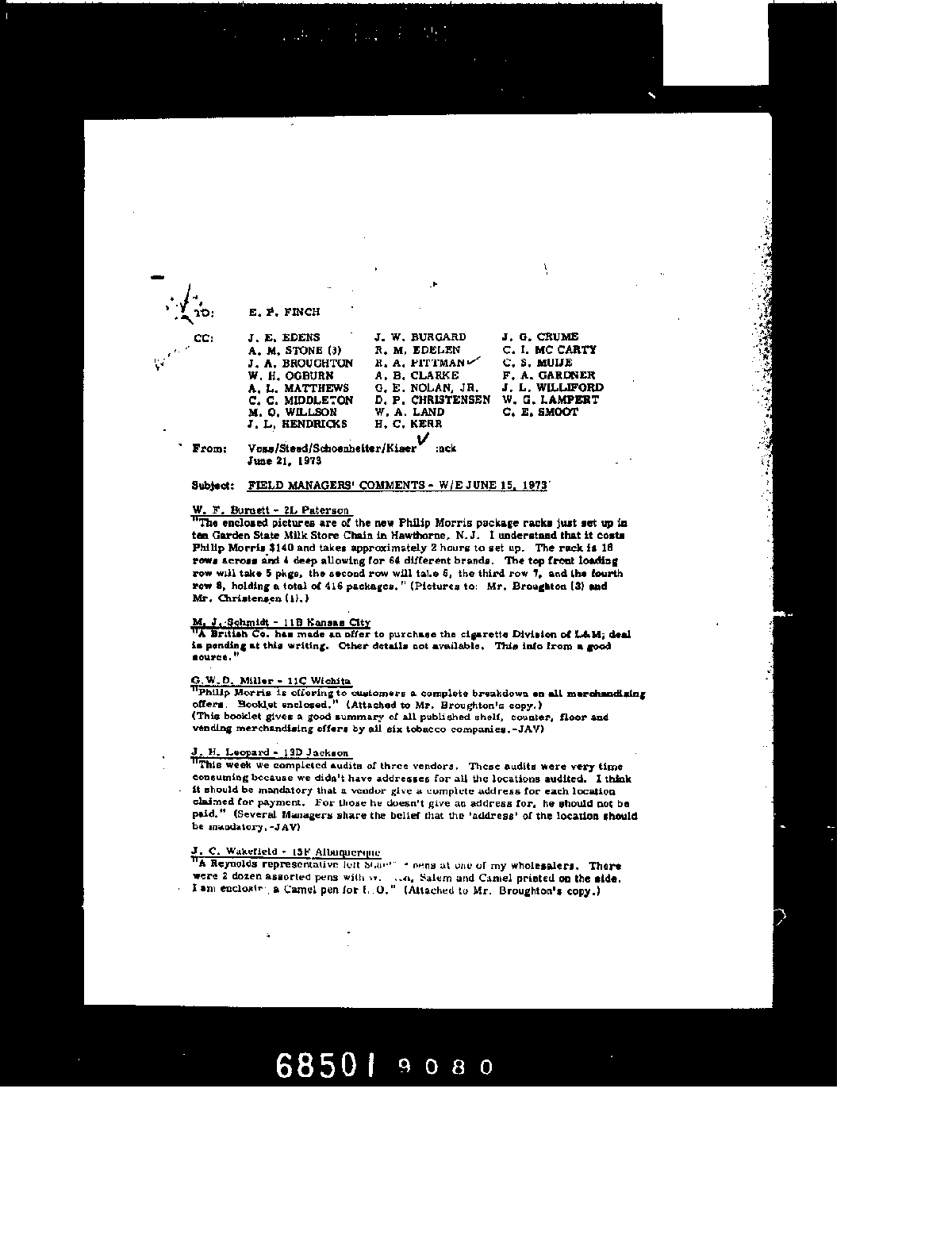}}
        \fbox{\includegraphics[width=0.24\linewidth,height=0.12\textheight]{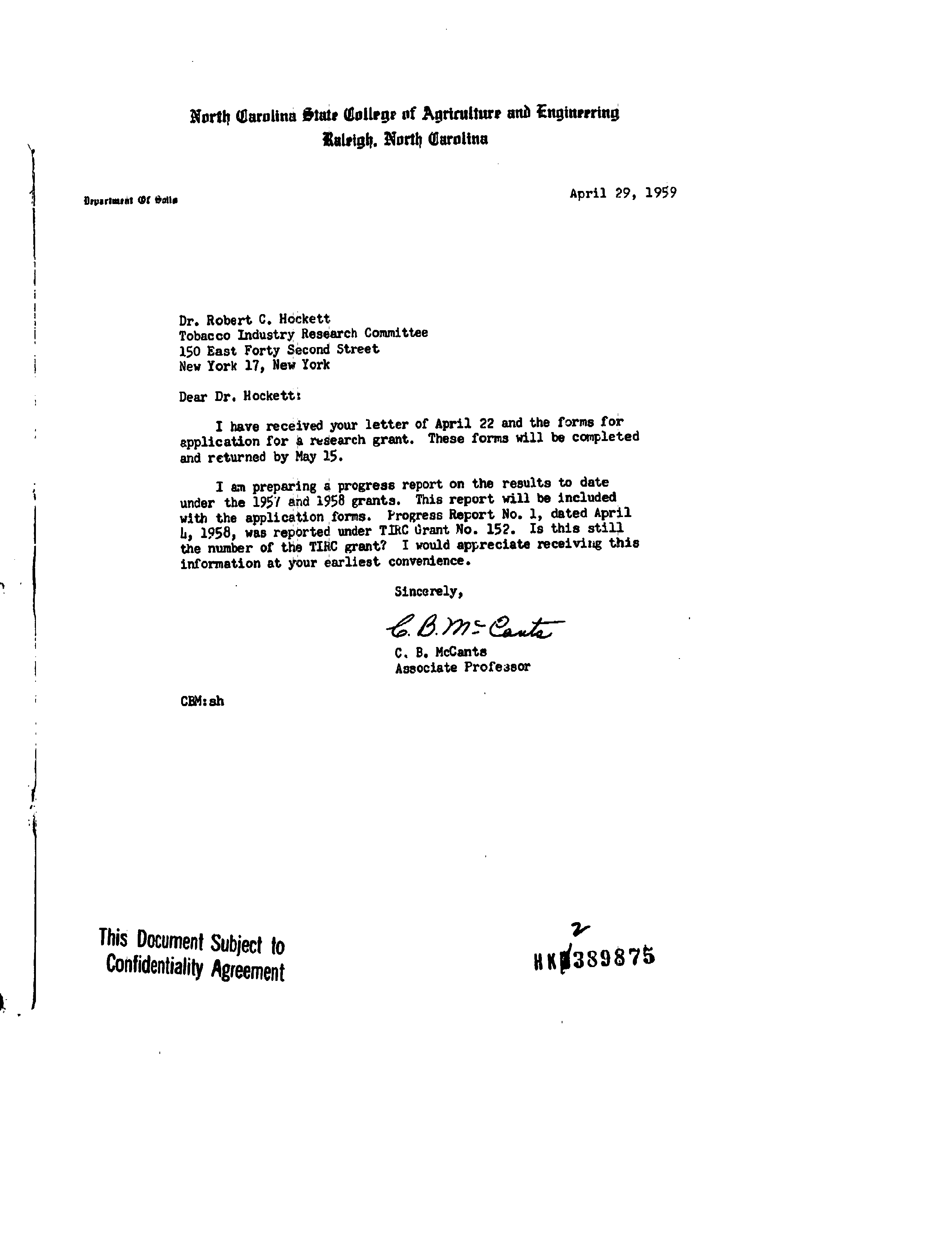}}
    \par\smallskip
        \fbox{\includegraphics[width=0.24\linewidth,height=0.12\textheight]{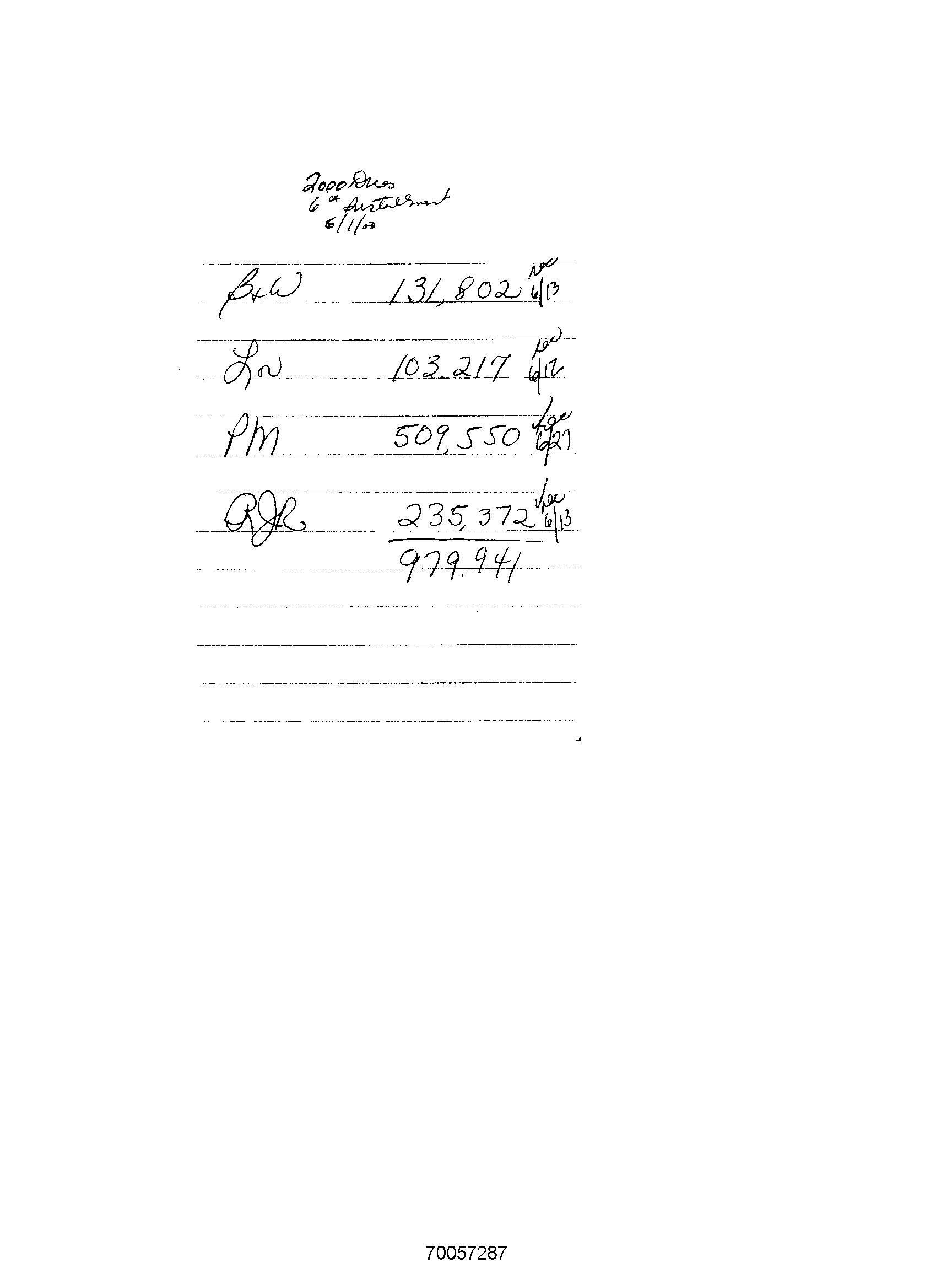}}
        \fbox{\includegraphics[width=0.24\linewidth,height=0.12\textheight]{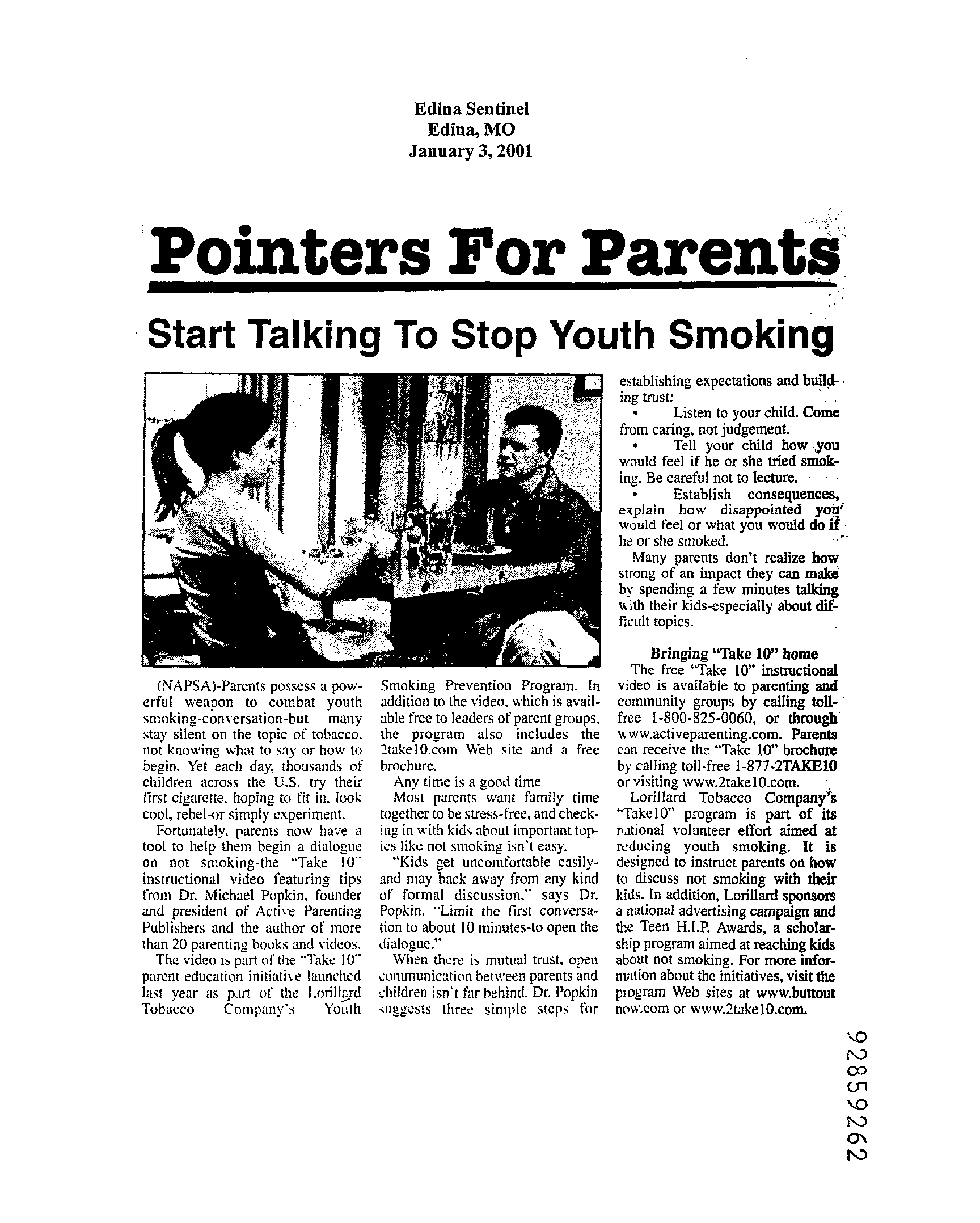}}
        \fbox{\includegraphics[width=0.24\linewidth,height=0.12\textheight]{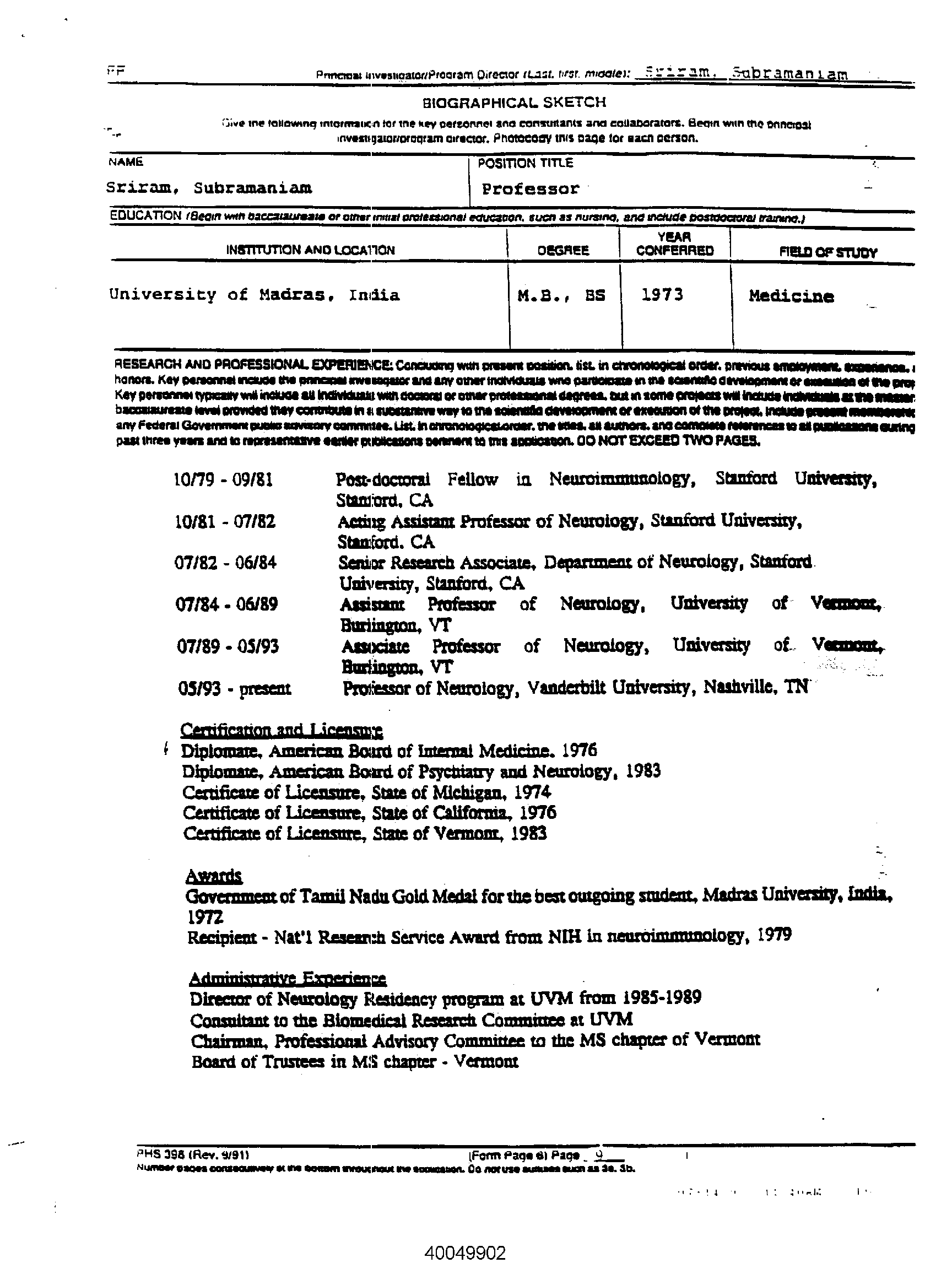}}
    \caption{Sample images from different classes of the Tobacco-3482 dataset.}
    \label{fig:tobacco-3482-overview}
\end{figure}

Today, business documents (cf. Fig.~\ref{fig:tobacco-3482-overview}) are often processed by document analysis systems (DAS) to reduce the human effort in scheduling them to the right person or in extracting the information from them.
One important task of a DAS is the classification of documents, i.e. to determine which kind of business process the document refers to. Typical document classes are \emph{invoice}, \emph{change of address} or \emph{claim} etc.
Document classification approaches can be grouped in image-based~\cite{afzal2015deepdocclassifier, lekang-14-a, harley2015icdar, doclass_Kumar12, doclass_Chen12, doclass_Kochi99, doclass_umamaheswara08} and content (OCR) based approaches~\cite{tang2016bayesian, diab2017using, li1998classification} (See Section~\ref{sec:related}.
DAS often include both variants. Which approach is more suitable often depends on the documents that are processed by the user. Free-form documents like usual letters normally need content-based classification whereas forms that contain the same text in different layouts can be distinguished by image-based approaches.

\begin{figure}
    \centering
        \fbox{\includegraphics[width=0.26\linewidth,height=0.13\textheight]{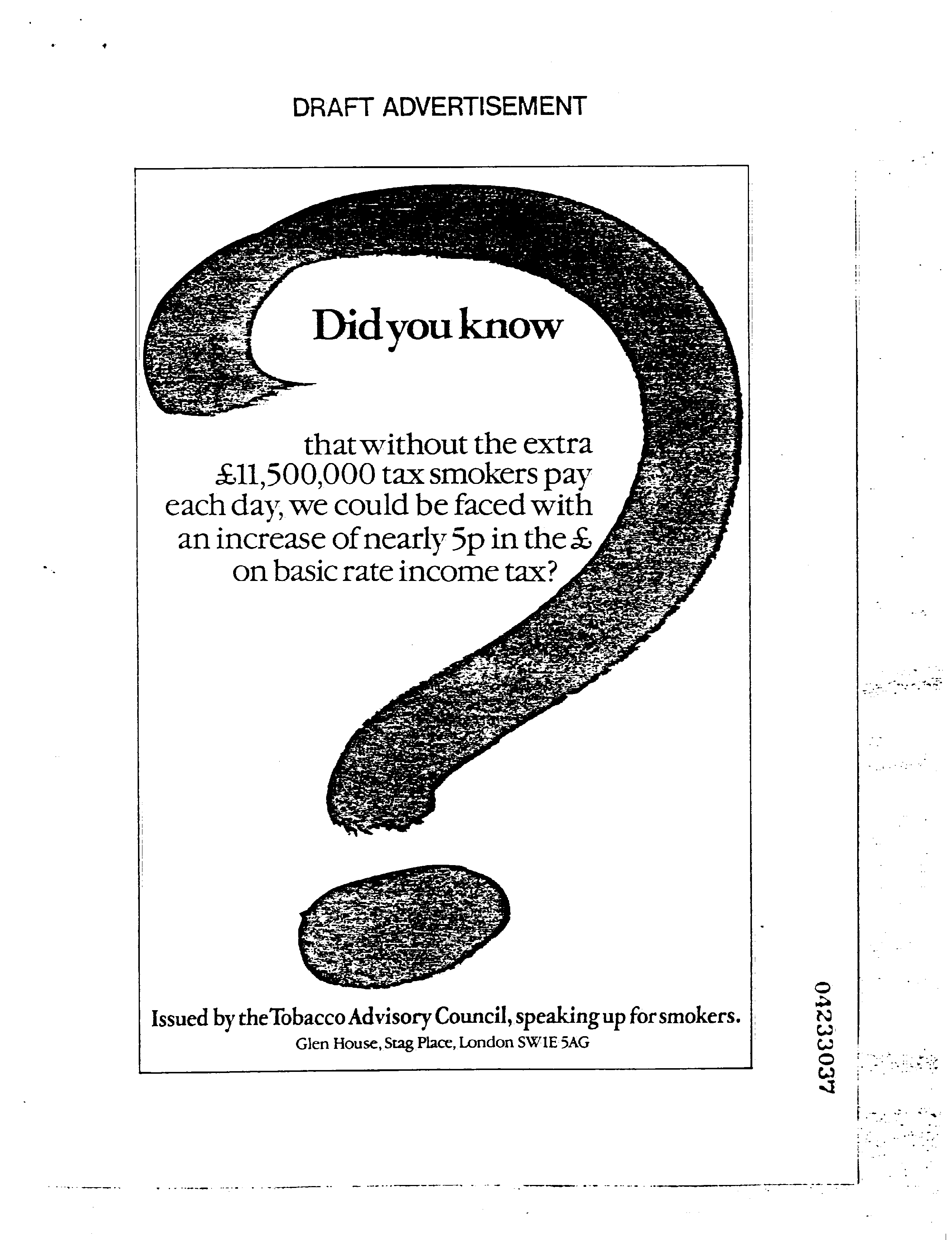}}
        \fbox{\includegraphics[width=0.26\linewidth,height=0.13\textheight]{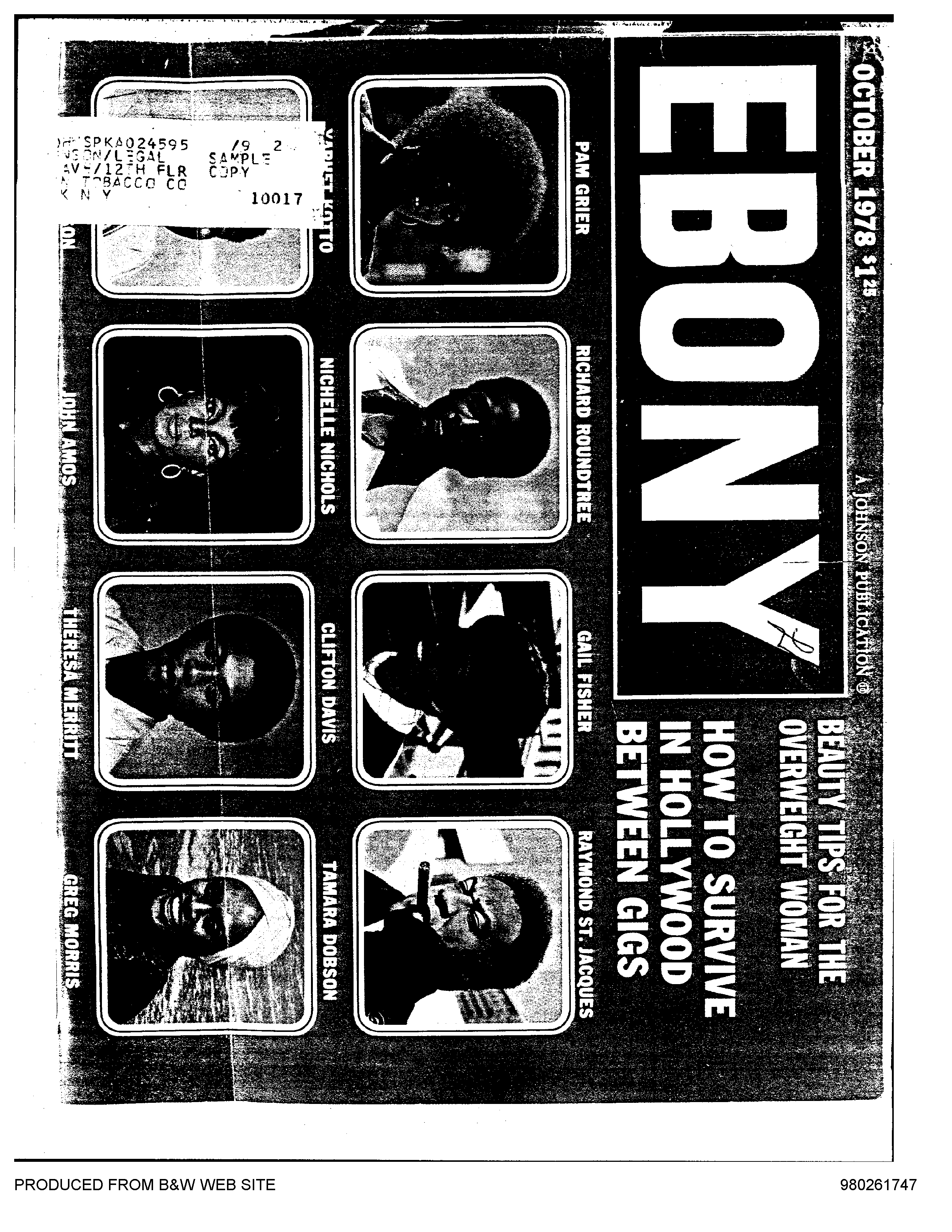}}
        \fbox{\includegraphics[width=0.26\linewidth,height=0.13\textheight]{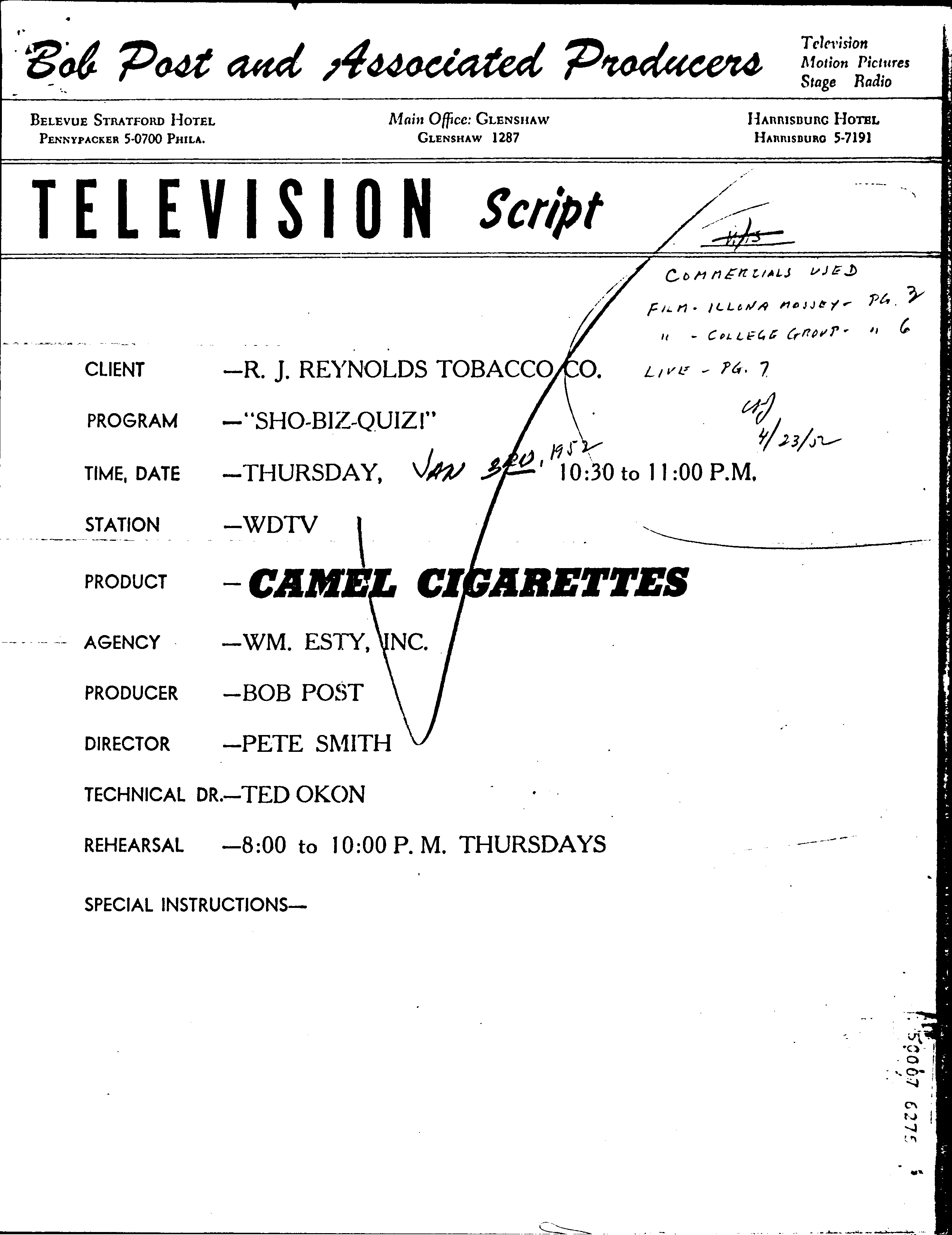}}
    \par\smallskip
        \fbox{\includegraphics[width=0.26\linewidth,height=0.13\textheight]{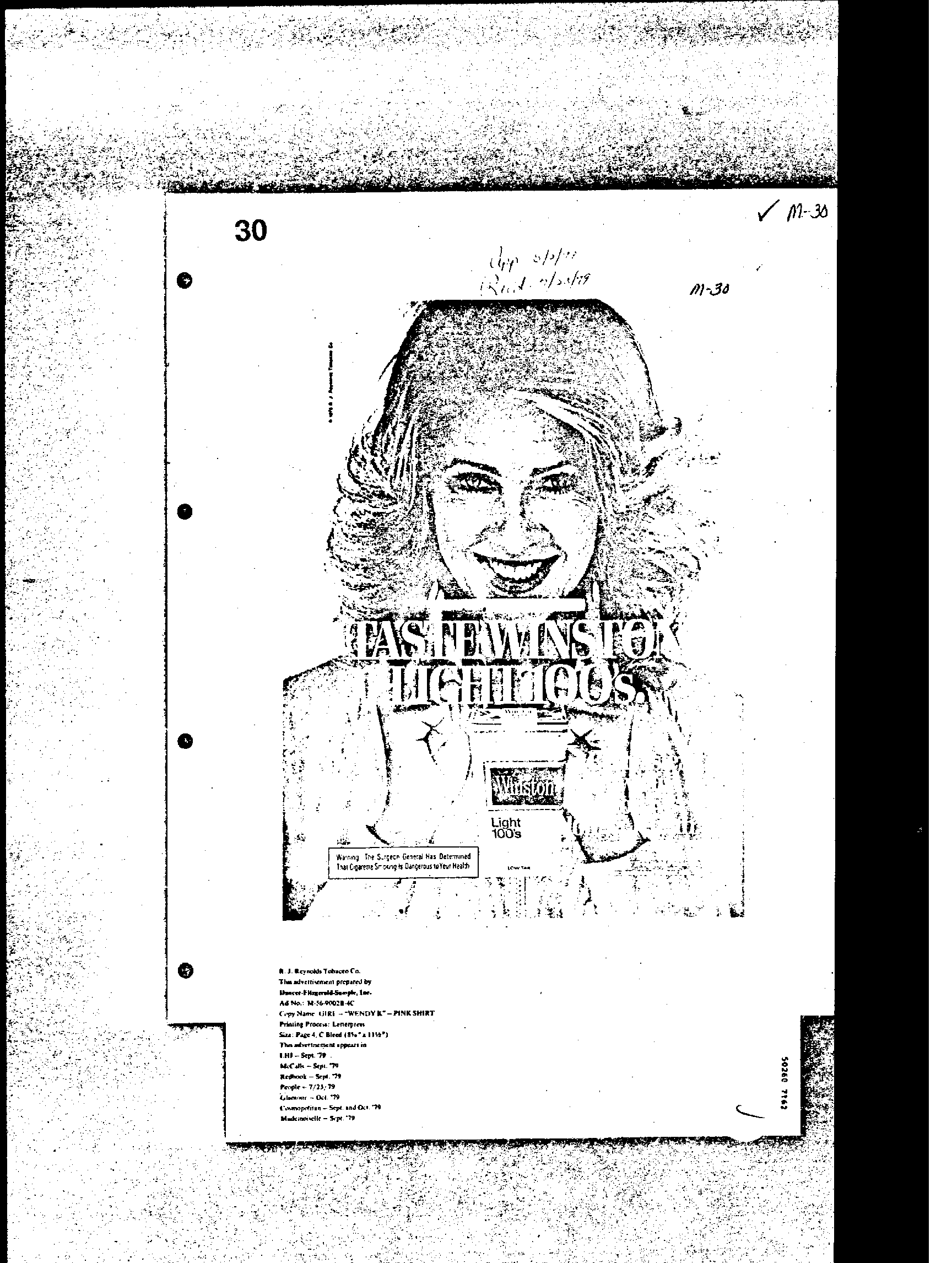}}
        \fbox{\includegraphics[width=0.26\linewidth,height=0.13\textheight]{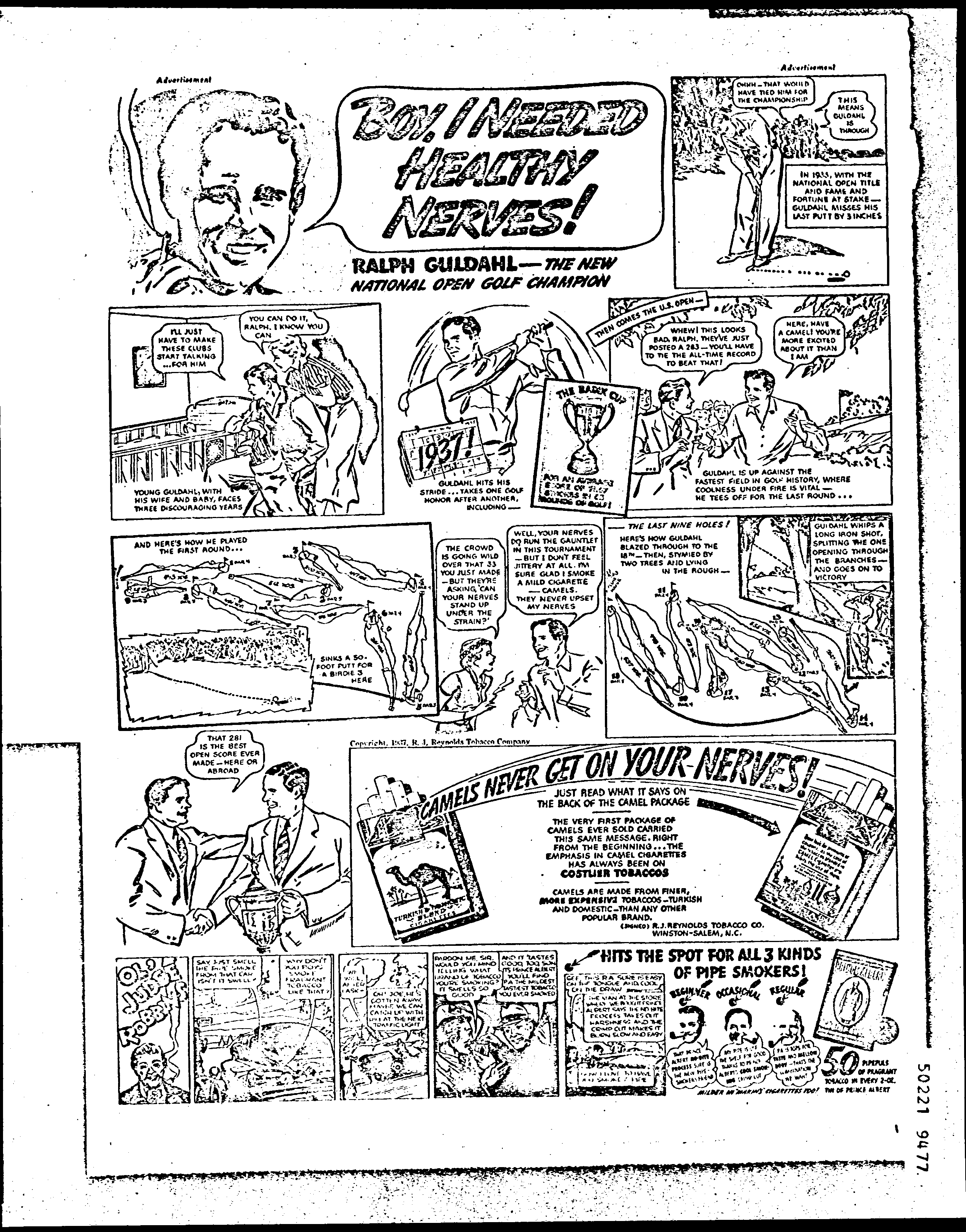}}
        \fbox{\includegraphics[width=0.26\linewidth,height=0.13\textheight]{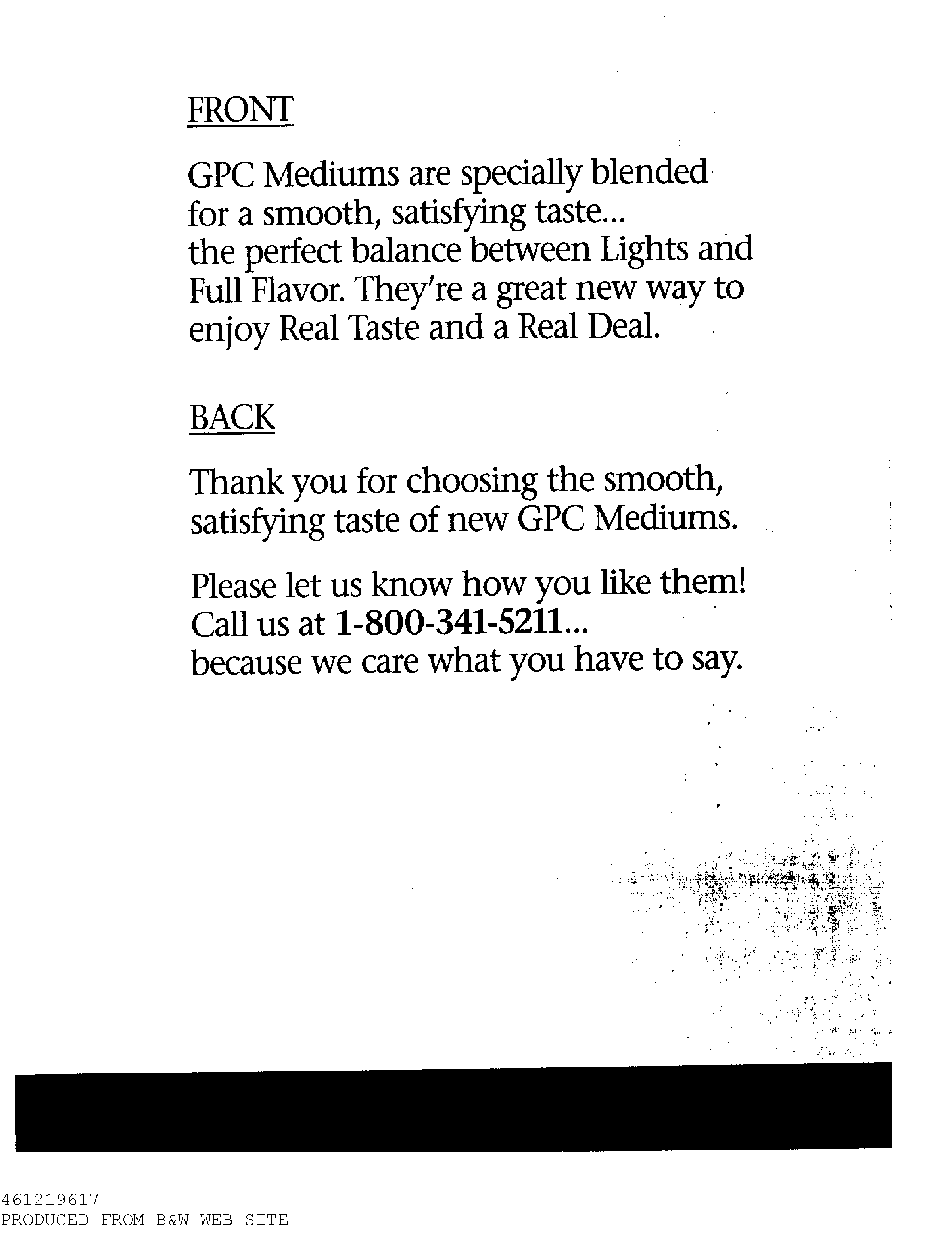}}
    \caption{Documents from the \emph{Advertisement} class of the Tobacco-3482 dataset showing a high intra-class variance.}
    \label{fig:tobacco-3482-intra}
\end{figure}

However, it is not always known in advance what category the document belongs to. That is why it is difficult to choose between image-based and content-based methods.
In general, the image-based approach is preferred that works directly on digitized images.
Due to the diversity of the document image classes, there exist classes with a high intra-class and low inter-class variance which is shown in Fig.~\ref{fig:tobacco-3482-intra} and Fig.~\ref{fig:tobacco-3482-inter} respectively. 
Hence it is difficult to come up with handcrafted features that are generic for document image classification.

With the increasing performance of convolutional neural networks (CNN) during the last years, it is more straightforward to classify images directly without extracting handcrafted features from segmented objects~\cite{afzal2015deepdocclassifier, lekang-14-a, harley2015icdar}.
However, these approaches are time-consuming at least during the training process. This means that it may take hours before the user gets feedback if the chosen approach for classification works in his case.
In addition, self-learning DAS that train incrementally based on the user's feedback will not have a good user experience because it just takes too long until the system improves while working with it.
The question is if there is an image-based approach for document classification which is efficient in classification and training as well.

In this paper, we propose to use \ac{elm}s which provide real-time training.
In order to overcome both, the hassle of manual feature extraction and long time of training, we devise a two-stage process that combines automatic feature learning of deep \ac{cnn}s with efficient \ac{elm}s.
The first phase is the training of a deep neural network that will be used as feature extractor.
In the second phase, \ac{elm}s are employed for the final classification.
\ac{elm}s are different in their nature from other neural networks (see Section~\ref{sec:elm}).
The presented work in the paper shows that it takes a millisecond on average to train over one image, hence showing a real-time performance.
This fact makes these networks also well-suited for usage in an incremental learning framework.

The rest of the paper is organized as follows: Section~\ref{sec:related} describes the related work in the field of document classification. A theoretical background on \ac{elm}s is given in Section~\ref{sec:elm}. In Section~\ref{sec:dcnn} the proposed combination of a deep \ac{cnn} and an \ac{elm} is described in detail. Section~\ref{sec:experiments} explains how the experiments are performed and presents the results. Section~\ref{sec:conclusion} concludes the paper and gives perspectives for future work.


%% file: relatedwork.tex
\section{Related Work}
\label{sec:related}

In the last years, a variety of methods has been proposed for document image classification. 
These methods can be grouped into three categories. 
The first category utilizes the layout/structural similarity of the document images. It is time-consuming to first extract the basic document components and then use them for classification.
The work in the second category is focused on the developing of local and/or global image descriptors. These descriptors are then used for document classification. 
Extracting local and global features is also a fairly time-consuming process. 
Lastly, the methods from the third category use \ac{cnn}s to automatically learn and extract features from the document images which are then classified. 
Nevertheless, also in this approach, the training process is very time-consuming, even using GPUs.
In the following, we give a brief overview of closely related approaches belonging to the three categories mentioned above.

\begin{figure}
    \centering
        \fbox{\includegraphics[width=0.26\linewidth,height=0.13\textheight]{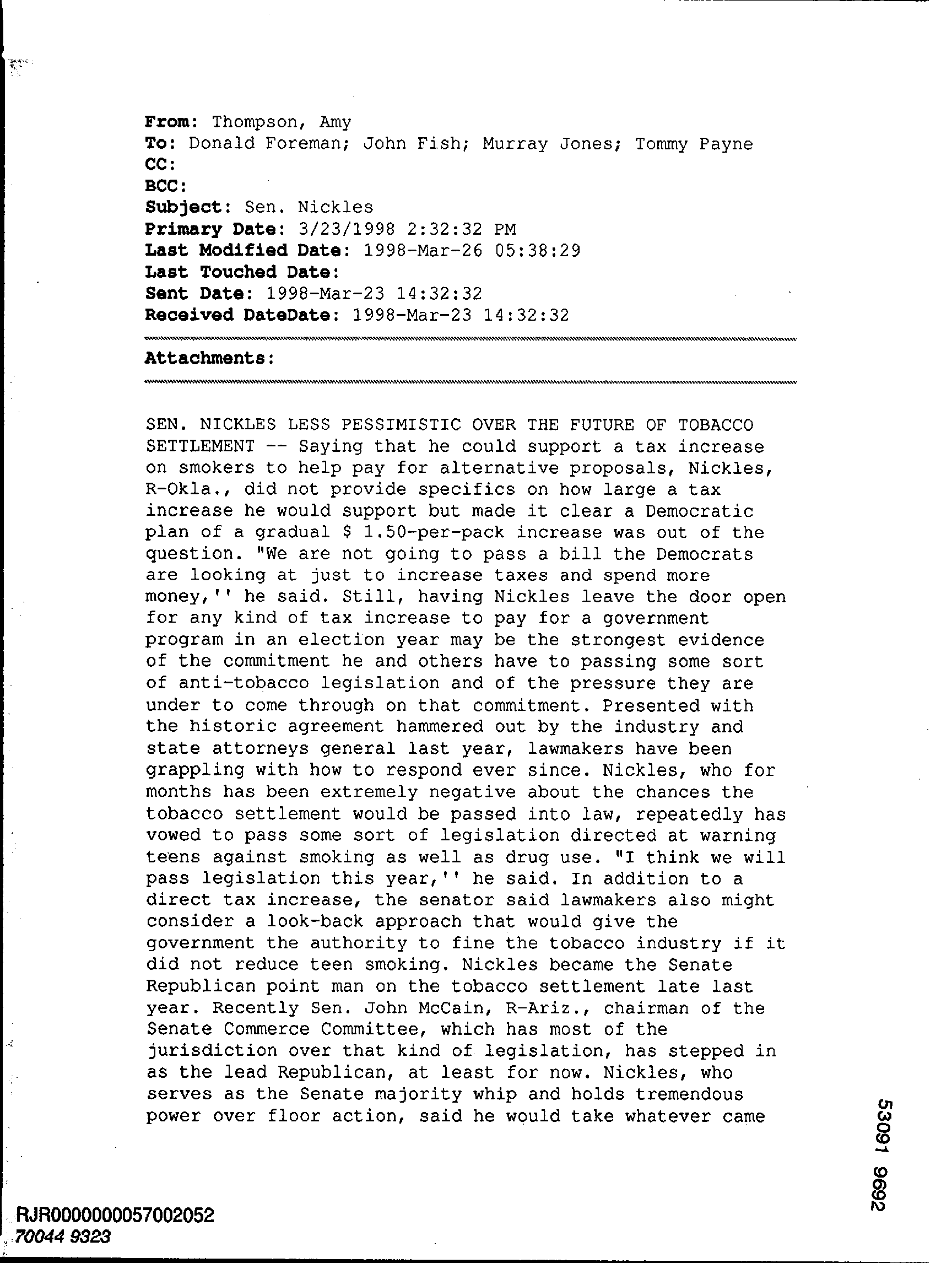}}
        \fbox{\includegraphics[width=0.26\linewidth,height=0.13\textheight]{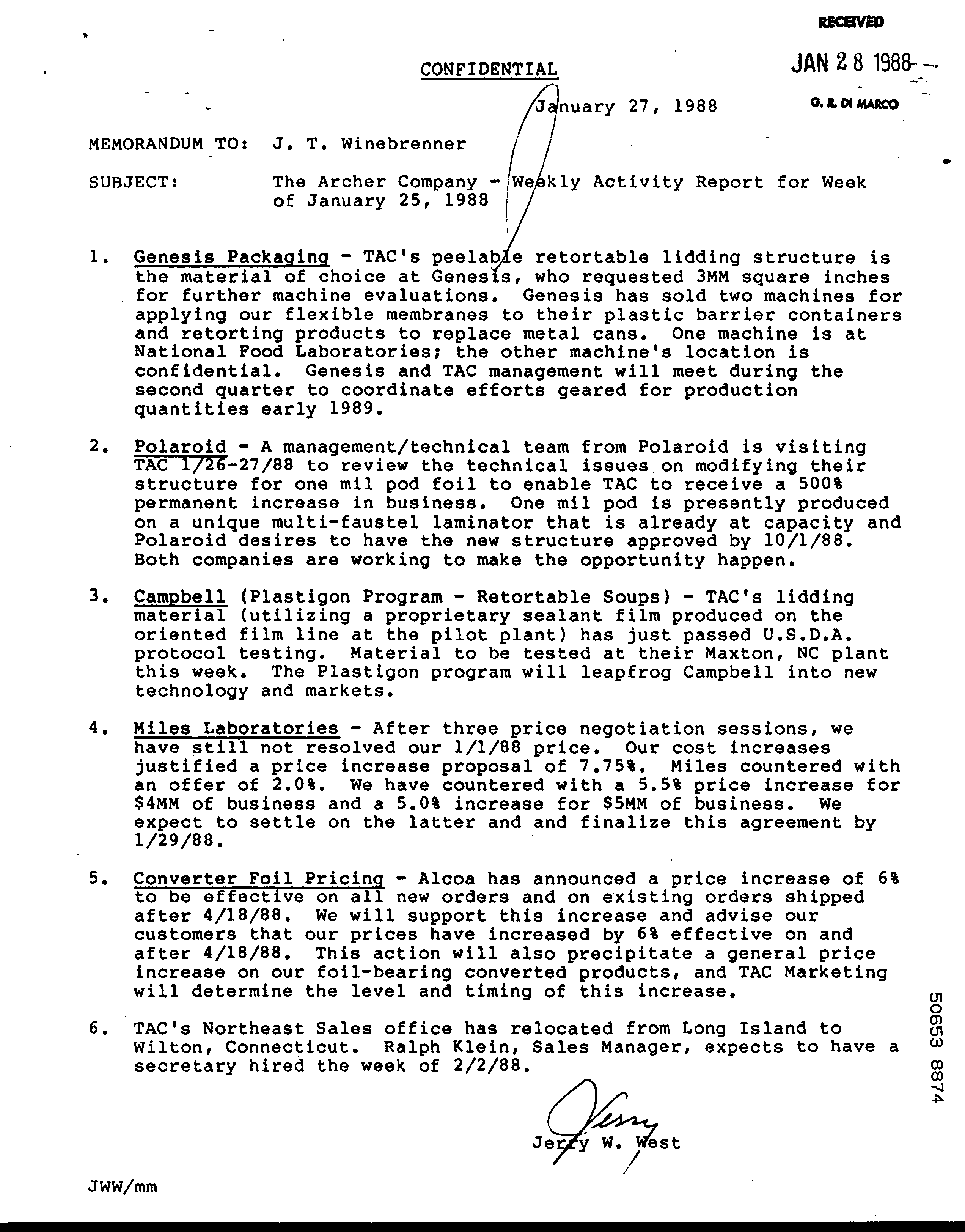}}
        \fbox{\includegraphics[width=0.26\linewidth,height=0.13\textheight]{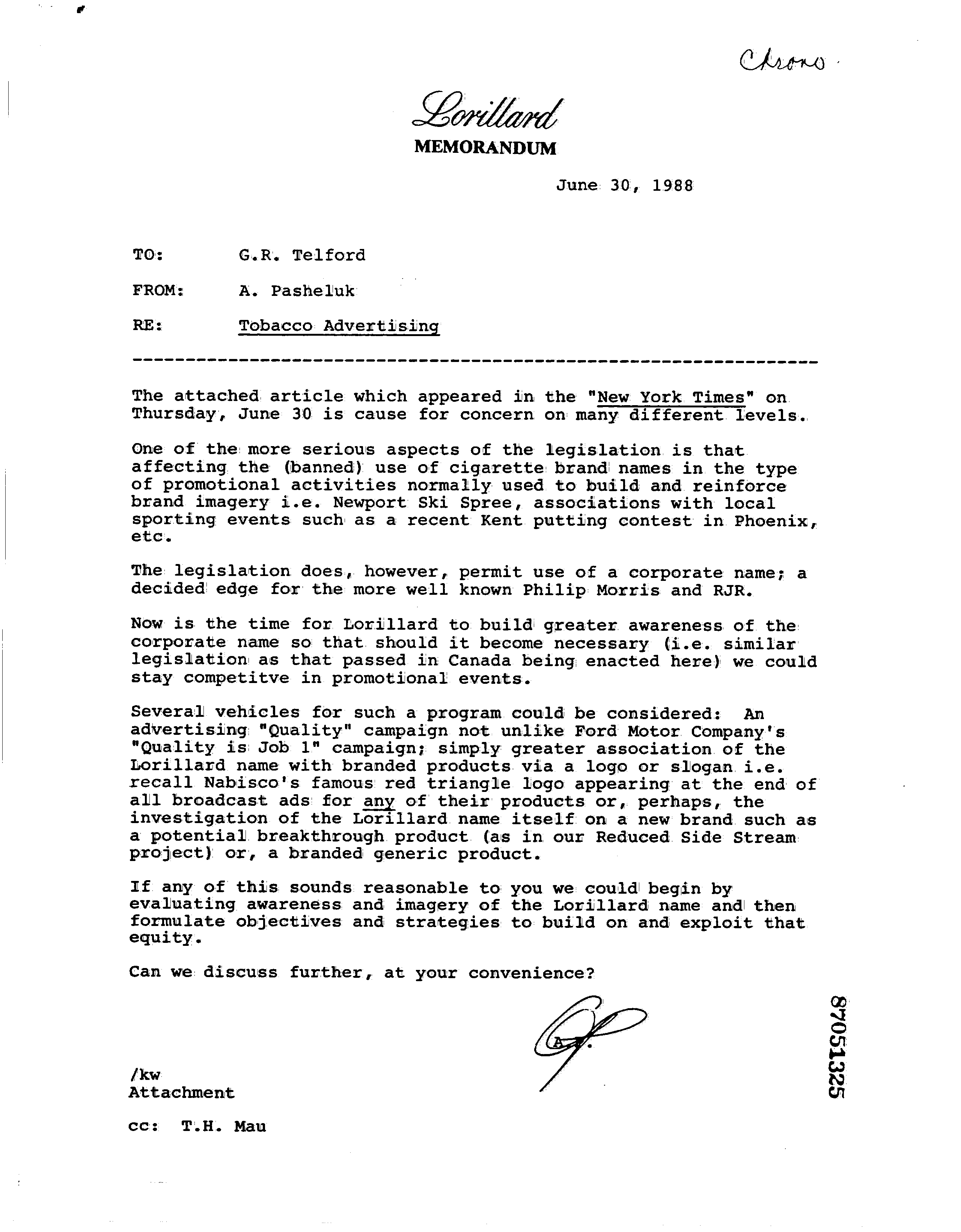}}
    \par\smallskip
        \fbox{\includegraphics[width=0.26\linewidth,height=0.13\textheight]{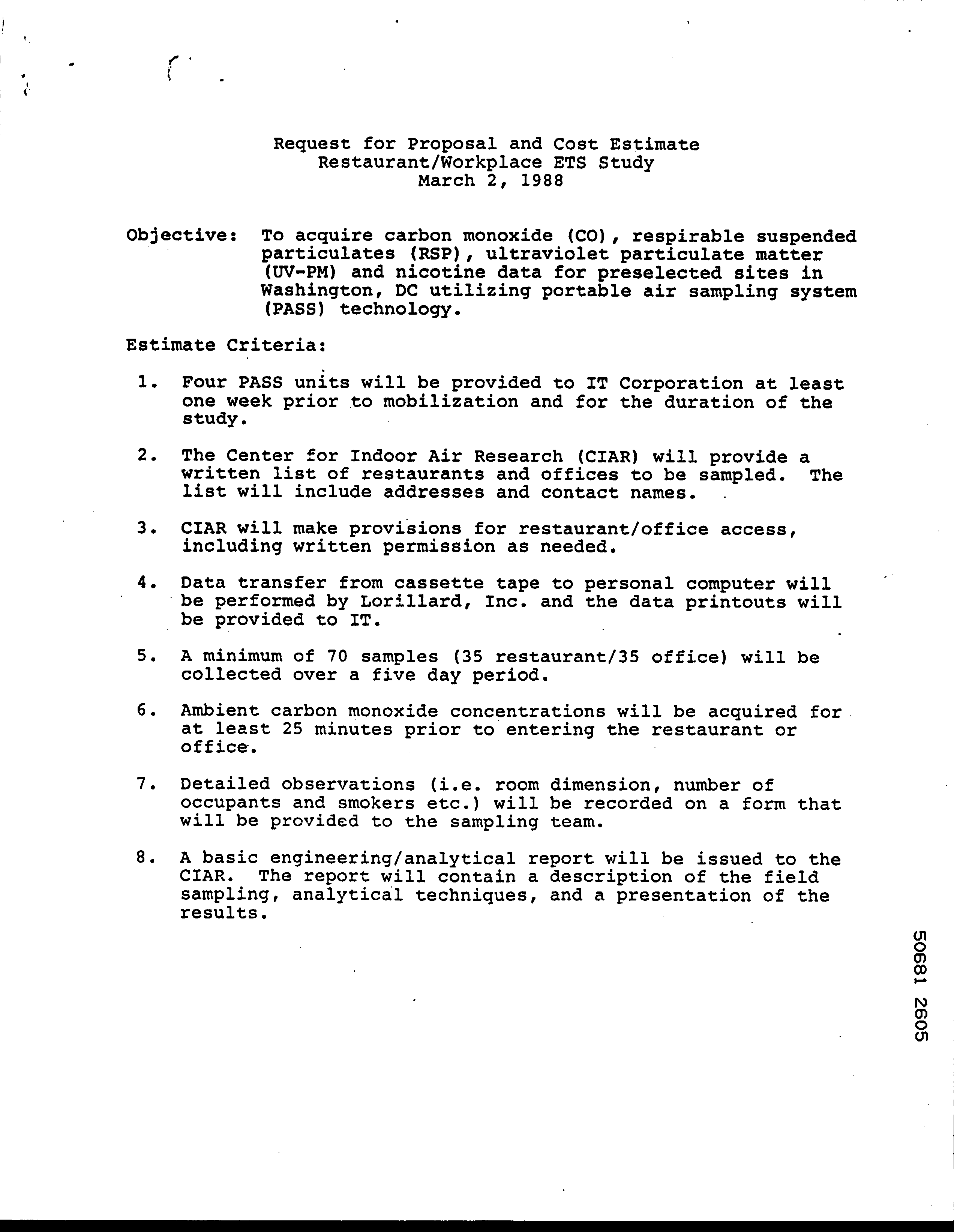}}
        \fbox{\includegraphics[width=0.26\linewidth,height=0.13\textheight]{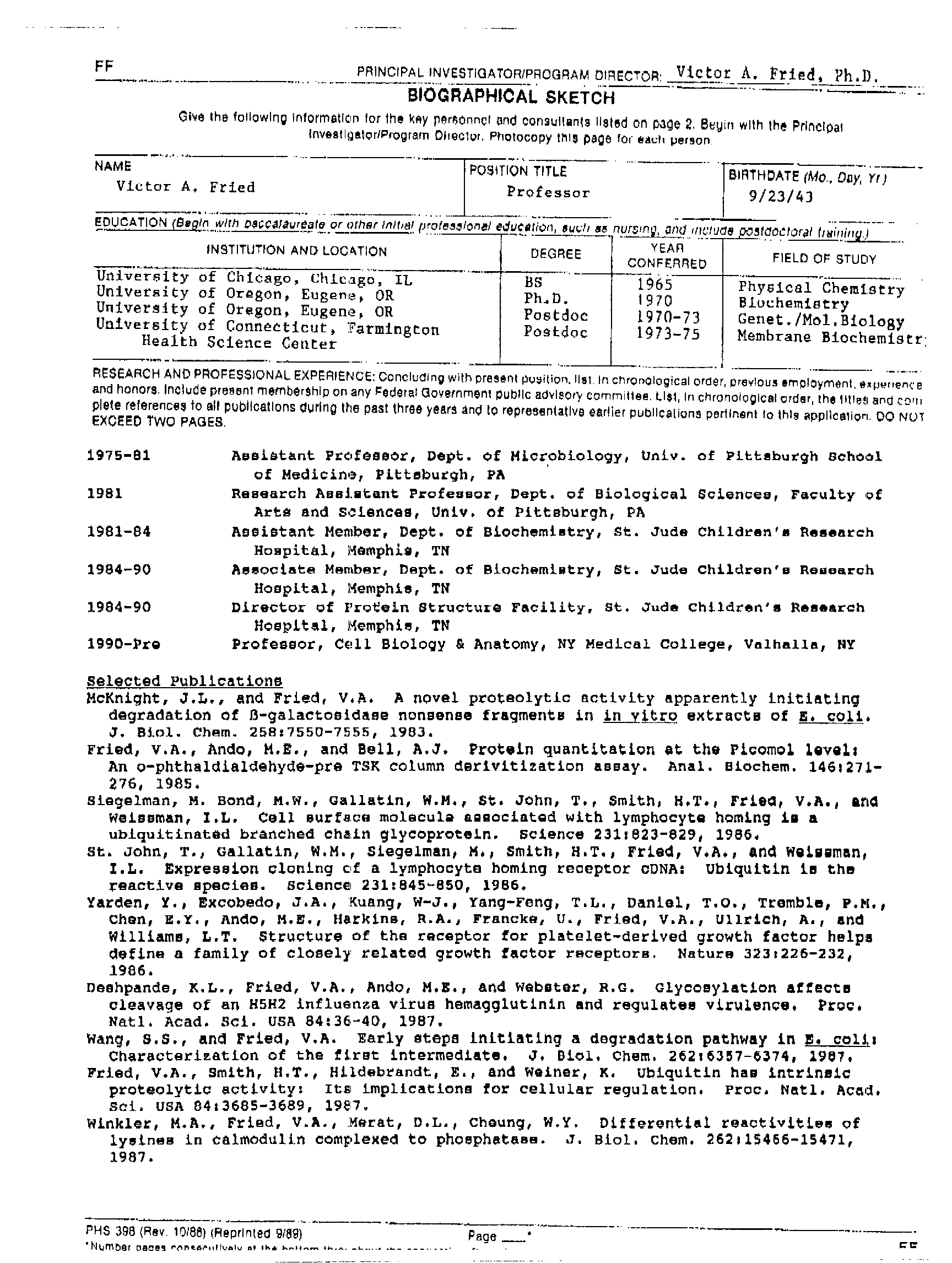}}
        \fbox{\includegraphics[width=0.26\linewidth,height=0.13\textheight]{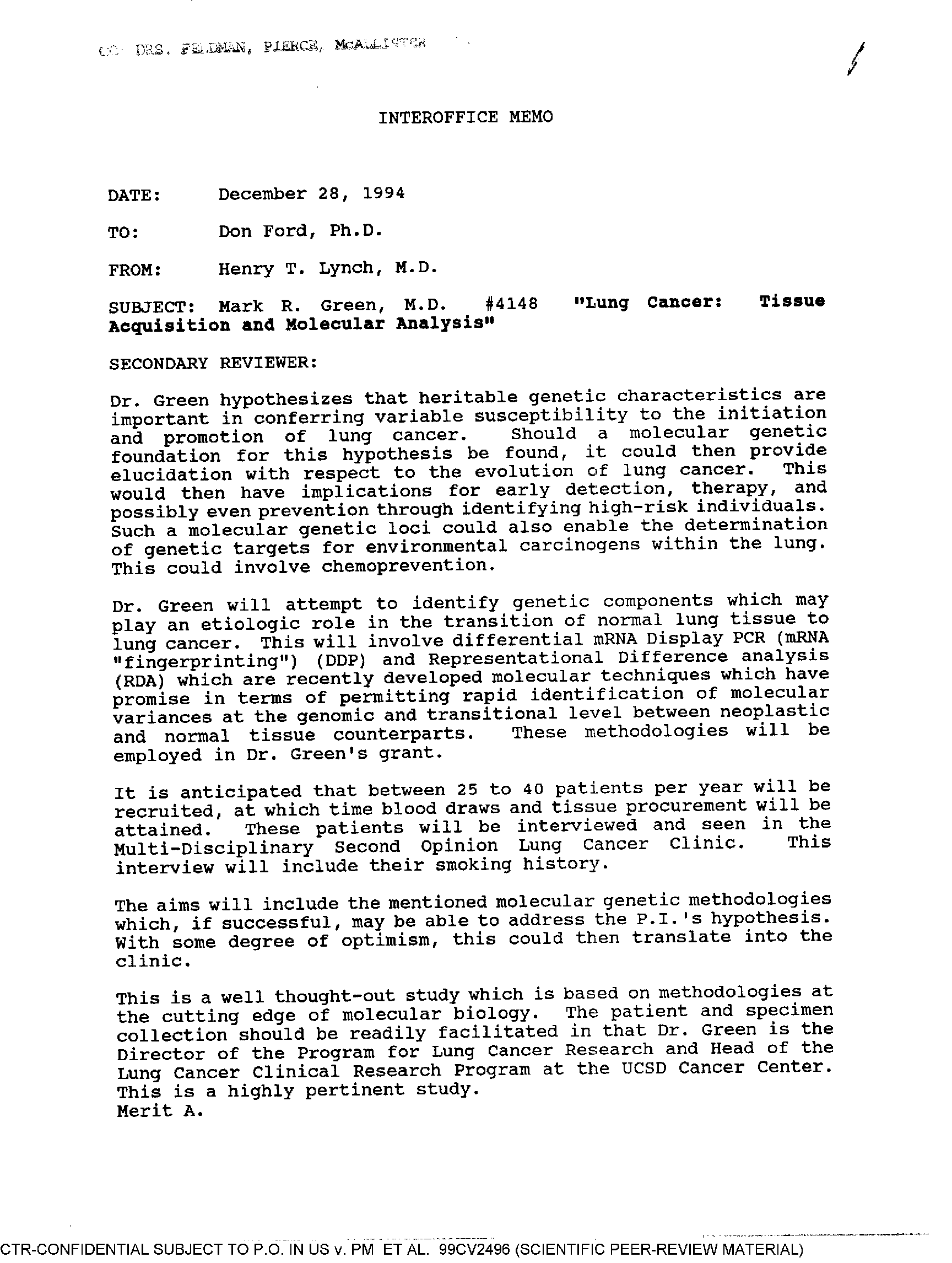}}
    \caption{Documents from different classes (\emph{Email}, \emph{Letter}, \emph{Memo}, \emph{Report}, \emph{Resume} and \emph{Scientific}) of the Tobacco-3482 dataset showing a low inter-class variance.}
    \label{fig:tobacco-3482-inter}
\end{figure}

Dengel and Dubiel~\cite{doclass_Dengel95} used decision trees to map the layout structure of printed letters into a complementary logical structure.
Bagdanov and Worring~\cite{doclass_Bagdanov2001} present a classification method for machine-printed documents that uses Attributed Relational Graphs (ARGs).
Byun and Lee~\cite{doclass_Byun2000} and Shin and Doermann~\cite{doclass_shin} used layout and structural similarity methods for document matching whereas, Kevyn and Nickolov~\cite{Collins-thompson02aclustering-based} combined both text and layout based features.

In 2012, Jayant et al.~\cite{doclass_Kumar12} proposed a method for document classification that relies on codewords derived from patches of the document images. The code book is learned in an unsupervised way on the documents. To do that, the approach recursively partitions the image into patches and models the spatial relationships between the patches using histograms of patch-codewords.
Two years later, the same authors presented another method which builds a codebook of SURF descriptors of the document images~\cite{doclass_Kumar14}. In a similar way as in their first paper, these features are then used for classification.
Chen et al.~\cite{doclass_Chen12} proposed a method which uses low-level image features to classify documents. However, their approach is limited to structured documents.
Kochi and Saitoh~\cite{doclass_Kochi99} presented a method that relies on pre-defined knowledge on the document classes. The approach uses models for each class of documents and classifies new documents based on their similarity to the models.
Reddy and Govindaraju~\cite{doclass_umamaheswara08} used pixel information from binary images for the classification of form documents. Their method uses the k-means algorithm to classify the images based on their pixel density.


Most important for this work are the \ac{cnn} based approaches by Kang et al.~\cite{lekang-14-a}, Harley et al.~\cite{harley2015icdar} and Afzal et al.~\cite{afzal2015deepdocclassifier}.
Kang et al. have been the first who used  \ac{cnn}s for document classification. Even though they used a shallow network due to limited training data, their approach outperformed structural similarity based methods on the Tobacco-3482 dataset~\cite{lekang-14-a}.
Afzal et al.~\cite{afzal2015deepdocclassifier}  and Harley et al. ~\cite{harley2015icdar} showed a great improvement in the accuracy by applying transfer learning from the domain of real-world images to the domain of document images, thus making it possible to use deep \ac{cnn} architectures even with limited training data. With their approach, they significantly outperformed the \sota at that time.
Furthermore, Harley et al. ~\cite{harley2015icdar} introduced the RVL-CDIP dataset which provides a large-scale dataset for document classification and allows for training \ac{cnn}s from scratch.

While deep \ac{cnn} based approaches have advanced significantly in the last years and are the current \sota, the training of these networks is very time-consuming. 
The approach presented in this paper belongs to the third category, but overcomes the issue of long training time.
To allow for real-time training while using the \sota performance of deep \ac{cnn}s, our approach uses a combination \ac{cnn}s~\cite{cnn_alexnet_nips2014} and \ac{elm}s~\cite{huang2004extreme,huang2006extreme}.

%% file: elm.tex
\section{Extreme Learning Machines}
\label{sec:elm}

\ac{elm} is an algorithm that is used to train \ac{slfn}~\cite{huang2006extreme, huang2004extreme}. The major idea behind \ac{elm} is mimicking the biological behaviour. While general neural network training uses backpropagation to adjust parameters i.e. weights, this step is not required for \ac{elm}s.
An ELM learns by updating weights in two distinct but sequential stages. These stages are random feature mapping and least square fitting. In the first stage, the weights between the input and the hidden layers are randomly initialized. In the second stage a linear least square optimization is performed and therefore no backpropagation is required.
The point that distinguishes \ac{elm} from other learning algorithms is the mapping of input features into a random space followed by learning in that stage.

In a supervised learning setting each input sample has a corresponding class label. Let $x$ and $t$ be the input sample and corresponding label respectively.

Let $X$ and $T$ be the sets of $n$ examples and represented as follows $\{X,T\} = \{x_k,t_k\}_{k=1}^{n}$ where $x_{k} \in {\mathbb{R}}^d$ and $t_{k} \in \mathbb{R}^m$ are the $k^{th}$ input and target vectors of $d$ and $m$ dimensions respectively.
The supervised classification searches for a function that maps the input vector to the target vector.
While there are many sophisticated forms of such functions~\cite{kotsiantis2007supervised}, one simple and effective function is single hidden layer feed-forward network (SLFN).
With respect to the setting described above a single layer network with $N$ hidden nodes can be depicted as follows

\begin{equation}
    o_j = \sum_{i=1}^{N} \beta_i g\left ( w_i^T x_j + b_i \right )
\end{equation}

where $w_i$ is the weight matrix connecting the $i_th$ hidden node and the input nodes, $\beta_i$ is the weight vector that connects the $i$th node to the output and $b_i$ is the bias. The function $g$ represents an activation function that could be $relu$, $sigmoid$, etc.

The above was the description of \ac{slfn}s. For \ac{elm}s the weights between the input and the hidden nodes $\left \{w_i,b_i  \right \}_{i=1}^N$ are randomly initialized. In the second stage, the parameters connecting the hidden and the output layer are optimized using regularized linear least square. Let $\psi(x_j)$ be the response vector from hidden layer to input $x_j$ and $B$ be the output parameter connecting the the hidden and output layer.
\ac{elm} minimizes the following sum of the squared losses.

\begin{equation}
\label{eq:elm}
 \frac{C}{2} \sum_{j=1}^{N} \left \| e_j \right \|_2^2 + \frac{1}{2}\left \|  \boldsymbol{B} \right \|_F^2 
\end{equation}
    
The second term in Eq.~\ref{eq:elm} is the regularizer to avoid the overfitting and $C$ is the trade-off coefficient.
By concatenating $\boldsymbol {H} = \left [ \psi(x_1)]^T,\psi(x_2)]^T \cdots \psi(x_N)]^T \right ]^T$ and $\boldsymbol{T} = \left [ t_1, t_2, \cdots t_N \right ]$ we get the following well known optimization problem called ridge regression.


\begin{equation}
\min_{ \boldsymbol{B} \in R^{N \times q }}  \frac{1}{2}\left \| \boldsymbol{B} \right \|_F^2 +  \frac{C}{2}  \left \|  \boldsymbol{T}- \boldsymbol{HB} \right \|_2^2
\end{equation}

The above mentioned problem is convex and constrained by the following linear system
\begin{equation}
\boldsymbol{B} + C \boldsymbol{H}^T \left ( \boldsymbol{T} - \boldsymbol{H}\boldsymbol{B} \right )=0
\end{equation}

This linear system could be solved using numerical methods for obtaining optimal $\boldsymbol{B}^*$

\begin{equation}
\boldsymbol{B}^*= \left ( \boldsymbol{H}^T \boldsymbol{H} + \frac{I_{N}}{C}\right )^{-1} \boldsymbol{H} \boldsymbol{T}
\end{equation}

%% file: deepcnn.tex
\section{Deep CNN and ELM for Document Image Classification}
\label{sec:dcnn}

\begin{figure*}
        \centering
        \includegraphics[width=0.85\textwidth]{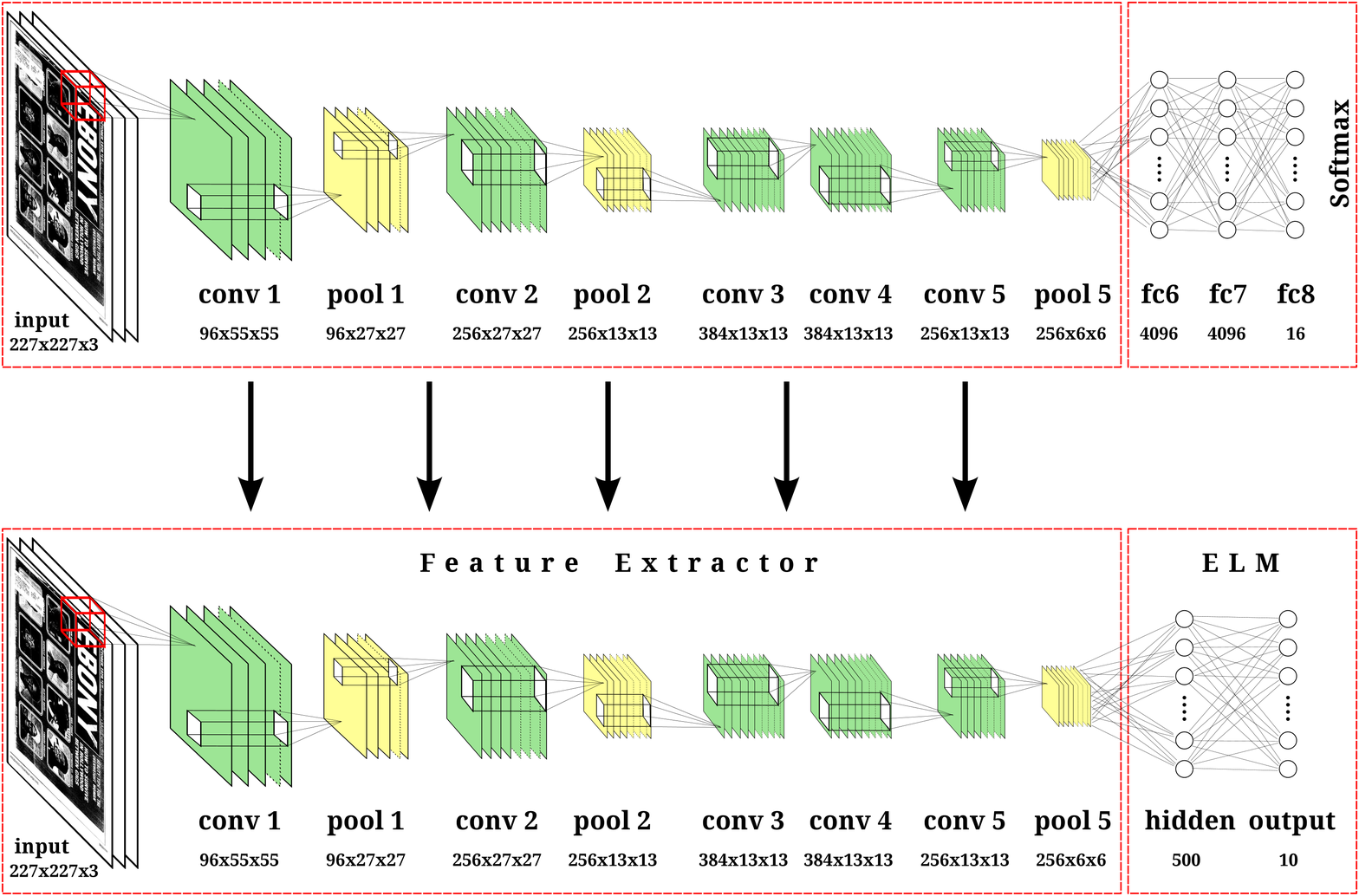}
        \caption{AlexNet is pretrained on the large scale dataset RVL-CDIP. Then, the fully-connected layers are replaced by an \ac{elm} and the other trained layers are copied to the new architecture.}
\label{fig:deepcnn}
\end{figure*}

This section presents in detail the mixed \ac{cnn} and \ac{elm} architecture and the learning methodology of the developed classification system.

\subsection{Preprocessing}
The method presented in this paper does not utilize document features that require a high resolution, such as optical character recognition. Instead, it solely relies on the structure and layout of the input documents to classify them. Therefore, in a preprocessing step, the high-resolution images are downscaled to a lower resolution of $227\times227$ which is the input size of the \ac{cnn}.

The common approach to successfully train \ac{cnn}s for object recognition is to augment the training data by resizing the images to a larger size and to then randomly crop areas from these images~\cite{cnn_alexnet_nips2014}. This data augmentation technique has proven to be effective for networks trained on the ImageNet dataset where the most discriminating elements of the images are typically located close to the center of the image and therefore contained in all crops. However, by this technique, the network is effectively presented with less than $80\,\%$ of the original image. We intentionally do not augment our training data in this way, because 
in document classification, the most discriminating parts of document images often reside in the outer regions of the document, \eg the head of a letter.

As a second preprocessing step, we subtract the mean values of the training images from both the training and the validation images.

Lastly, we convert the grayscale document images to RGB images, \ie we copy the values of the single-channel images to generate three-channel images. 



\subsection{Network Architecture}

The deep CNN architecture proposed in this paper is based on the AlexNet architecture \cite{cnn_alexnet_nips2014}. It consists of five convolutional layers which are followed by an Extreme Learning Machine (ELM) \cite{huang2006extreme}.


As in the original AlexNet architecture, we get $256$ feature maps of size $6\times6$ after the last max-pooling layer (\cf~Fig.~\ref{fig:deepcnn}).

While AlexNet uses multiple fully-connected layers to classify the generated feature maps, we propose to use a single-layer \ac{elm}.

The weights of the convolutional layers are pretrained on a large dataset as a full AlexNet, \ie with three subsequent fully-connected layers and standard backpropagation.
After the training has converged, the fully-connected layers are discarded and the convolutional layers are fixed to work as a feature extractor.
The feature vectors extracted by the \ac{cnn} stub then provide the input vectors for the \ac{elm} training and testing (\cf~Fig.~\ref{fig:deepcnn}).

The \ac{elm}s used in this architecture is a single-layer feed-forward neural network. We test \ac{elm}s with $2000$ neurons in the hidden layer and 10 output neurons, as the target dataset has 10 classes. The neurons use sigmoid as activation function.

\subsection{Training Details}
As already stated, we train a full AlexNet on a large dataset to provide a useful feature extractor for the \ac{elm} and then train the \ac{elm} on the target dataset. Specifically, we train AlexNet on a dataset, which contains images from $16$ classes. Therefore, the number of neurons in the last fully-connected layer of AlexNet is changed from $1,000$ to $16$.

All, but the last network layer are initialized with an AlexNet model\footnote{https://github.com/BVLC/caffe/tree/master/models/bvlc\_alexnet} that was pretrained on ImageNet.
The training is performed using stochastic gradient descent with a batch size of $25$, an initial learning rate of $0.001$, a momentum of $0.9$ and a weight decay of $0.0005$. To prevent overfitting, the sixth and seventh layers are configured to use a dropout ratio of $0.5$. After $40$ epochs, the training process is finished. The caffe framework~\cite{jia2014caffe} is used to train this model.

The \ac{elm}s are trained and evaluated on the Tobacco-3482 dataset \cite{doclass_Kumar14} which contains images from $10$ classes. The images are passed through the CNN stub and the activations of the fifth pooling layer are presented to the \ac{elm} (\cf~Fig.~\ref{fig:deepcnn}).detail

%% file: experiments.tex
\section{Experiments and Results}
\label{sec:experiments}

\begin{figure}
        \centering
        \includegraphics[width=\linewidth]{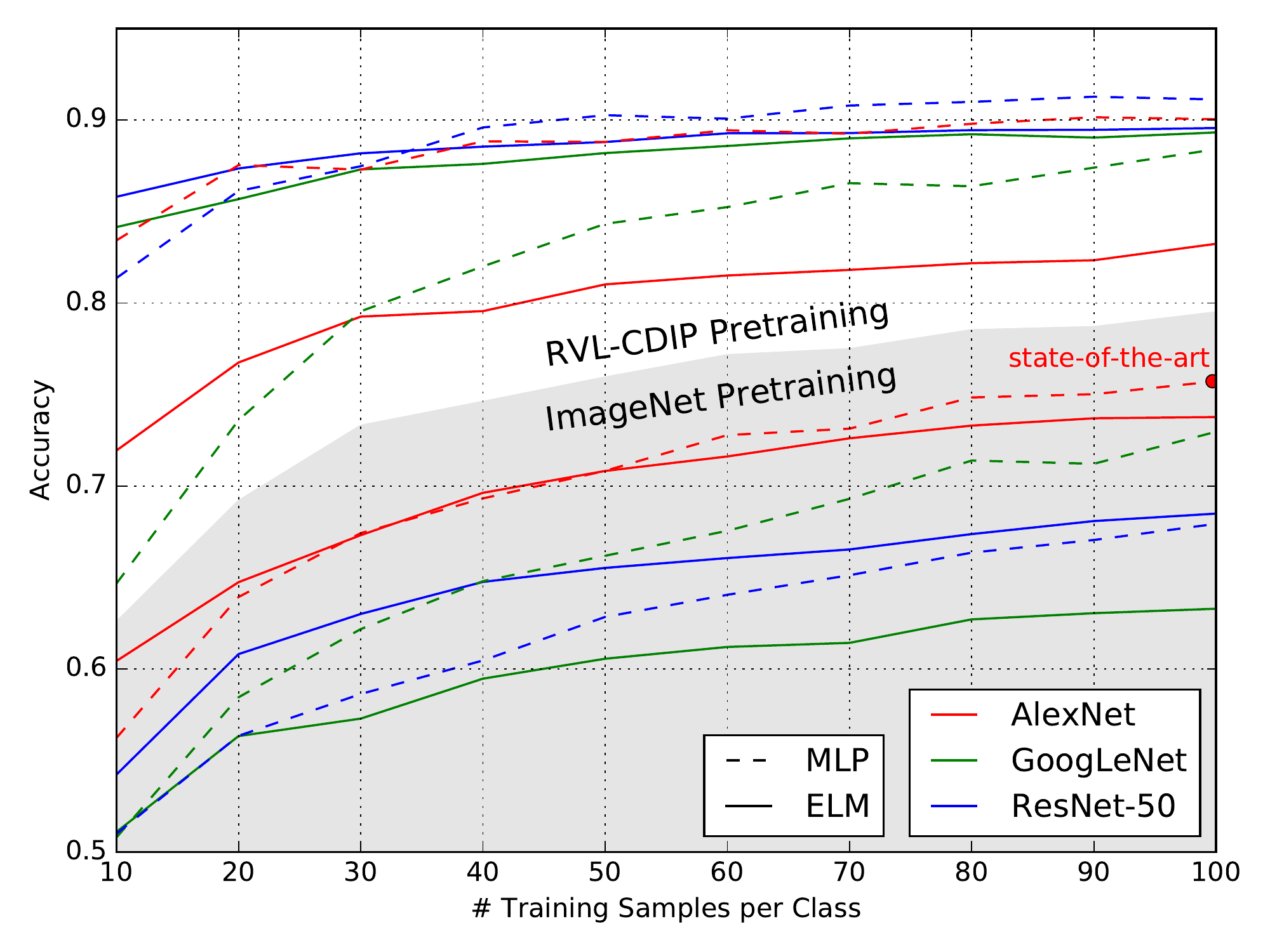}
        \caption{Mean accuracy achieved by the different \ac{elm} classifiers in comparison to the original networks.}
\label{fig:accuracy}
\end{figure}

\subsection{Datasets}

In this paper, two datasets are used. First, we use the Ryerson Vision Lab Complex Document Information Processing (RVL-CDIP) dataset \cite{harley2015icdar} to train a full AlexNet. This dataset contains $400,000$ images which are evenly distributed across $16$ classes. $320,000$ of the images are dedicated for training, $40,000$ images are each dedicated for validation and testing.

Secondly, we use the Tobacco-3482 dataset \cite{doclass_Kumar14} to train the presented ELM and evaluate its performance. This dataset contains $3,482$ images from ten document classes. 

As there exists some overlap between the two datasets, we exclude the images that are contained in both datasets from the large dataset. Therefore, AlexNet is not trained on $320,000$ but only on $319,784$ images.

\subsection{Evaluation Scheme}
\label{sec:eval}
To allow for a fair comparison with other approaches on the Tobacco-3482 dataset, we use a similar evaluation protocol as Kang et al. \cite{lekang-14-a} and Harley et al. \cite{harley2015icdar}. Specifically, we conduct several experiments with different training datasets. We only use subsets of the Tobacco-3482 dataset for training ranging from $10$ images per class to $100$ images per class. The remaining images are used for testing. Since the dataset is so small, for each of these dataset splits, we randomly create ten different partitions to train and evaluate our classifiers and report the median performance.
Note, that the ELMs are not optimized on a validation set. Thus, there is no validation set needed.

\begin{table}
\renewcommand{\arraystretch}{1.3}
\centering
\caption{Accuracy achieved on the Tobacco-3482 dataset by the different classifiers with different pretraining. Here, all networks use $100$ images per class during training and the rest for testing. The reported accuracy is the mean accuracy achieved on 10 different dataset partitions.}
\begin{tabular}{l|c}
 & $\emptyset$ Accuracy \\\hline
Structural methods~\cite{afzal2015deepdocclassifier} & $40.3\,\%$ \\\hline
AlexNet (ImageNet) & $75.73\,\%$ \\\hline
AlexNet-ELM (ImageNet) & $73.77\,\%$ \\\hline
AlexNet (RVL-CDIP) & $90.05\,\%$ \\\hline
AlexNet-ELM (RVL-CDIP) & $83.24\,\%$

\end{tabular}
\label{tab:accuracy}
\end{table}

\begin{table}
\renewcommand{\arraystretch}{1.3}
\centering
\caption{Time needed to train and test the classifiers using a NVidia Tesla K20x as GPU and an Intel i7-6700K @ 4.00GHz as CPU. The testing time is the time required to classifiy the entire test set of $2482$ images.}
\begin{tabular}{l|c|c}
 & Training & Testing \\\hline
AlexNet (GPU) & 10 min, 34 sec & 3480 ms \\\hline
AlexNet-ELM (GPU) & 1176 ms & 3066 ms \\\hline
AlexNet (CPU) & 6 h, 44 min, 8 sec & 4 min, 30 sec \\\hline
AlexNet-ELM (CPU) & 1 min, 26 sec & 4 min, 19 sec
\end{tabular}
\label{tab:runtime}
\end{table}

\subsection{Experiments}

As a first and baseline experiment, we train AlexNet which is pretrained on ImageNet, on the Tobacco-3482 dataset as was already done by Afzal et al.~\cite{afzal2015deepdocclassifier}. As described above, we train multiple versions of the network with $10$ different partitions per training data size. In total, 100 networks are trained, \ie $10$ networks on each $10$, $20$, ..., $100$ training images per class. The training datasets for these experiments are further subdivided into a dataset for actual training ($80\,\%$) and a dataset for validation ($20\,\%$) (\cf~\cite{harley2015icdar}).

Secondly, we train an ImageNet initialized AlexNet on $319,784$ images of the RVL-CDIP corpus and discard the fully-connected part of the network. The network stub is used as a feature extractor to train and test our ELMs. The ELMs are trained on the Tobacco-3482 dataset as described in section~\ref{sec:eval}. As these networks depend on random initialization, we train $10$ ELMs for each of the $100$ partitions and report the mean accuracy for each partition size.



\subsection{Results}

The performance of our proposed classifiers in comparison to the current \sota is depicted in Fig.~\ref{fig:accuracy}. As can be seen, the \ac{elm} classifier with document pretraining already outperforms the current state-of-the-art with as little as $20$ training samples per class.
With $100$ training samples per class, the test accuracy can be increased from $75.73\,\%$ to $83.24\,\%$ (\cf~Table~\ref{tab:accuracy}) which corresponds to an error reduction of more than $30\,\%$.

Together with the exceptional performance boost the runtime needed for both training and testing is reduced (\cf~Table~\ref{tab:runtime}). Especially in the case of GPU accelerated training, the proposed approach is more than $500$ times faster than the current \sota. For both training and testing, the combined \ac{cnn}/\ac{elm} approach needs about $1$ ms per image, thus making it real-time. As more than $90\,\%$ of the total runtime are used for the feature extraction, a different \ac{cnn} architecture could speed this up even further.

The \ac{elm} classifier with ImageNet pretraining achieves an accuracy which is comparable to that of the current state-of-the-art at a fraction of the computational costs.

Note, that AlexNet pretrained on the RVL-CDIP dataset and fine-tuned on the Tobacco-3482 dataset achieves even better performance in terms of accuracy. However, as this would be as slow as the current \sota, this is not in the scope of this paper. The main idea of this work is to provide a fast and accurate classifier.

A confusion matrix of an exemplary \ac{elm} classifier which was trained on $100$ images per class is shown in Fig.~\ref{fig:confusion}. As can be seen, the class \emph{Scientific} is by far the hardest to recognize. This result is consistent with Afzal et al.~\cite{afzal2015deepdocclassifier} and can be explained by low inter-class variance between the classes \emph{Scientific} and \emph{Report}.

\subsection{Experiments with deeper architectures}
For completeness, we also conduct the described experiments with GoogLeNet~\cite{szegedy2015going} and ResNet-50~\cite{he2016deep} as underlying network architectures. As depicted in Fig.~\ref{fig:accuracy}, the networks perform extremely well.

However, since both of these architectures have only one fully connected layer for classification which is replaced by the \ac{elm}, there is no runtime improvement at inference time, but only at training time.
Furthermore, due to the depth of these models, we have to drastically reduce the batch size which decreases the degree of parallelism and makes these approaches not viable for real-time training with a single GPU.

%% file: conclusion.tex
\begin{figure}
        \centering
        \includegraphics[width=\linewidth]{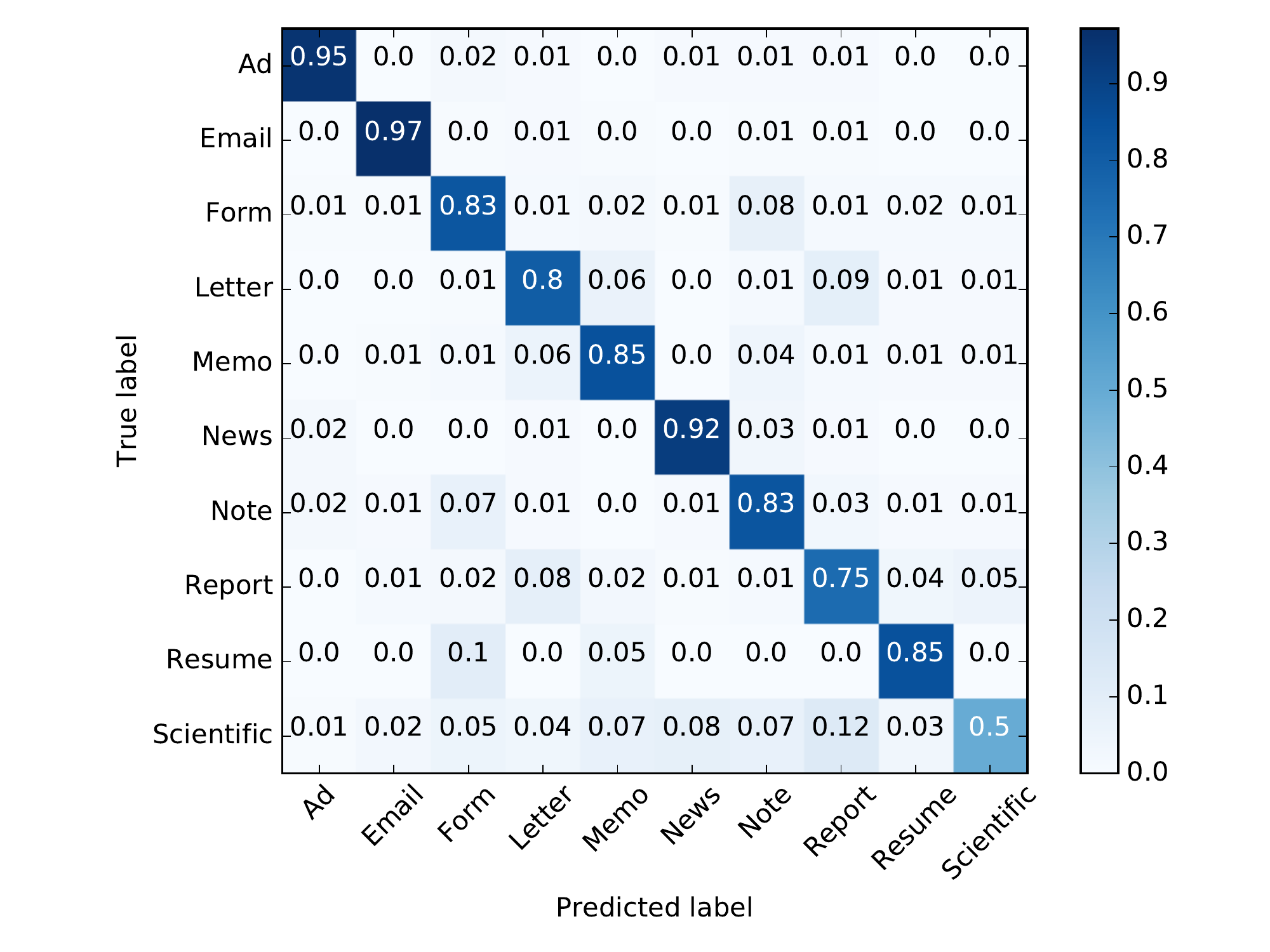}
        \caption{Confusion matrix of an exemplary ELM.}
\label{fig:confusion}
\end{figure}

\section{Conclusion and Future Work}
\label{sec:conclusion}
We have addressed the problem of real-time training for document image classification.
In particular, we present a document classification approach that trains in real-time, i.e. a millisecond per image and outperforms the current state-of-the-art by a large margin.
We suggest a two-stage approach which uses feature extraction from deep neural networks and efficient training using \ac{elm}.
The latter stage leads to superior performance in terms of efficiency.
Several quantitative evaluations show the power and potential of the proposed approach.
This is a big leap forward for DAS that are bound to quick system responses.

An interesting future dimension is the fast extraction of image features, because, in the presented approach over $90$\,\% of the time is consumed for feature extraction from deep neural networks.
Another future experiment is to benchmark the GoogLeNet and ResNet-50 based \ac{elm} classifiers in a high-performance cluster.